1

# Title: Microscopic Robots That Sense, Think, Act, and Compute

**Authors:** Maya M. Lassiter†[1], Jungho Lee†[2], Kyle Skelil[3], Li Xu[2], Lucas Hanson[4], William H. Reinhardt[1], Dennis Sylvester[2], Mark Yim[3], David Blaauw[2], Marc Z. Miskin[1*]

**Affiliations:**

[1]Department of Electrical and Systems Engineering, University of Pennsylvania

[2]Department of Electrical Engineering and Computer Science, University of Michigan

[3]Department of Mechanical Engineering and Applied Mechanics, University of Pennsylvania

[4]Department of Physics and Astronomy, University of Pennsylvania

*Corresponding author. Email: mmiskin@seas.upenn.edu

† These authors contributed equally to this work.

**Abstract:** While miniaturization has been a goal in robotics for nearly 40 years, roboticists have struggled to access sub-millimeter dimensions without making sacrifices to on-board information processing due to the unique physics of the microscale. Consequently, microrobots often lack the key features that distinguish their macroscopic cousins from other machines, namely on-robot systems for decision making, sensing, feedback, and programmable computation. Here, we take up the challenge of building a microrobot comparable in size to a single-celled paramecium that can sense, think, and act using onboard systems for computation, sensing, memory, locomotion, and communication. Built massively in parallel with fully lithographic processing, these microrobots can execute digitally defined algorithms and autonomously change behavior in response to their surroundings. Combined, these results pave the way for general purpose microrobots that can be programmed many times in a simple setup, cost under $0.01 per machine, and work together to carry out tasks without supervision in uncertain environments.

**Summary:** We demonstrate autonomous sub-millimeter robots with on-board sensing, computation, memory, communication, and locomotion.

**Main Text:**

***Introduction:*** Natural microorganisms demonstrate the feasibility of building autonomous, intelligent systems at dimensions too small to see by eye, a fact that has fascinated roboticists for decades (*1*, *2*). Ideally, a microscopic robot would retain all the features that distinguish their macroscopic cousins from other machines: it would be able to sense, compute actions, support repeated programming, and manipulate or explore the surrounding world. Yet miniaturizing robots without forgoing many of these hallmarks has proven difficult.

Currently, the smallest robots with fully integrated systems for sensing, programmable computation, and motion control sit at or above a millimeter (*3*), a size first reached over twenty years ago (*4*). Further shrinking these early successes has been hampered by the fact that many of the physical laws that govern semiconductor circuits(*5*), energy storage(*6*), power transfer(*7*), and macroscale propulsion(*8*, *9*) scale super-linearly with size, compounding into significant problems as robots attempt to shrink under sub-millimeter dimensions. As a result, fundamentally different approaches are required for truly microscopic robots (*10*).

Typically, roboticists circumvent the barriers to reaching the microscale by externally controlling locomotion through peripheral hardware, sacrificing programmability, sensing and/or autonomy in the process (*9*, *11*–*13*). These devices, while highly functional, remain tied to external equipment to make decisions and execute tasks, frequently cannot sense or respond to their environment, and/or can only transition between a limited number of hardcoded behaviors on command. Without integrated information processing, microrobots struggle with changing or unknown environments and offer limited reconfigurability after fabrication, blunting their usefulness in real-world applications (*13*).

In this work, we take a different approach, directly addressing the challenge of building a robot that can sense its environment, compute digitally programmable actions, and change its behavior, all while fitting in a space that is too small to see with the naked eye. This advance reduces the volume of the smallest digitally programmable robot 10,000-fold and opens the door to microscopic robots that autonomously sense, think, and act. At the same time, moving computation to the microrobot reduces both the cost and operational overhead to a bare minimum, paving a new path to widespread adoption.

***Results*:** Our approach to microrobots leverages silicon semiconductor electronics to form the information processing systems. Remarkably, many of the essential components for a smart microrobot, notably small-scale sensors, actuators, and information processing systems have been demonstrated individually in packages a few hundred microns in size (*15*–*18*). Yet integrating all these systems into a single microrobot without violating constraints on power, size, manufacturability, or performance presents a series of unresolved challenges. First, microrobots inherently have limited power budgets. For electronically controlled versions, this means on-board circuits must be made in mature process technologies to reduce leakage across electrical components at the expense of transistor density. Second, any propulsion system controlled by the circuits must also be small and low power, yet simple enough to integrate on top of the complex topography of a commercially fabricated CMOS chip without damaging the underlying electronics. Finally, robots must be freely released at the end of fabrication with high-yield techniques that scale to large numbers of machines.

We address these challenges with a series of innovations spanning circuit design and nanofabrication. On the circuit side, we operate within a limited power budget of ~100 nW by building our robot in a 55 nm CMOS process and leveraging sub-threshold digital logic. This process's high threshold voltage reduces transistor leakage enough to support a variety of on-board circuits in a sub-mm package without violating power constraints. For instance, at this node, we can build full microprocessors since, when properly designed, the leakage for this circuit, which dominates the active power, can be below 15 nW. Indeed, Fig. 1A and 1B show that in a space just 210 x 340 x 50 $\mu m^3$ we can fit photovoltaic cells for power, sensors for temperature, four actuator control circuits, an optical receiver for downlink communication and programming, a processor, and memory. All these sub-circuits are optimized to minimize size and power consumption, fabricated at a commercial foundry, and detailed in our prior publication and/or methods section (*19*). We build two robot configurations that differ only in the number of photovoltaic cells, resulting in body widths of 210 and 270μm, though, as shown later, we find no significant differences between them in performance. At this scale, the robot's size and power budget are both comparable to many unicellular microorganisms(*20*) and Figure 1C shows the robot beside plant cells for scale.

*Computation and communication:* The fabrication processes used to achieve dense data storage in commercial electronics are incompatible with the current leakage requirements for an autonomous microrobot. As a result, data on a microrobot must be stored in larger, low-leakage transistor-based memory, which at this process node limits the robot's storage capacity to hundreds of bits. To achieve digitally controlled behaviors despite this limited memory size, the robot's processor runs a custom designed complex instruction set architecture (CISC-ISA). This design compresses useful robot actions into specialized processor instructions (e.g., sense the environment, move for N cycles, etc.) and critically is well suited for microrobots because it offloads memory footprint without sacrificing functionality (see Table 1 for a full list of commands). As a fully functional computer, the programmable memory defines both the sequences of instructions and the robot's states. Unlike prior work on electronic microrobots(*21*), the way that states are updated is also digitally reconfigurable.

The commands run by the processor can be split into two groups, one for manipulating data and the other for implementing robot-specific functions. To handle data, we include decode logic for the 11-bit CISC-ISA instruction set, a 32-entry 11-bit instruction memory, a register file with four 8-bit registers, and a 16-entry 8-bit data memory. The instruction set features conventional arithmetic operations such as addition (add), subtraction (sub), bitwise AND/OR (and/or), and shift left/right (sll/srl). It also supports control flow operations including unconditional jumps and conditional branches based on the comparison results of two registers. Finally, data transfer operations perform inter-register transfers using a move (mov) command and data movement between registers and data memory with store (sb) and load (lb) commands.

The robot specific instructions interface on-board data with the physical world through sensors and actuators. They include a command for motion control (mot), which drives a user specified voltage sequence to the actuators N-times, a sensing instruction (ts) which measures the temperature around the robot and stores the value in memory, and a communication instruction (wav) which takes the value in a register, Manchester-encodes it, and then modulates selected actuators to output the data. Each of the robotic instructions requires multiple clock cycles to complete. However, to simplify programming and reduce program size, each halts instruction execution so that from the programmer's point of view they appear to complete in a single cycle. In effect, this design compresses into a single command what would otherwise require dozens of instructions, making it possible to perform useful tasks with minimal instruction memory.

To further reduce the required memory size, we use progressive, multi-step programming via the communication down-link. First, we load the microrobot with an initialization program to configure the desired actuator state. Second, we give task programs that define the robot's operation. This two-step approach uses the base-station as a temporary memory buffer, making the most of the robot's limited memory (roughly 500 bits).

We send both the initialization and task programs to robots using an optical communication link, shown in Fig. 1D. Two light emitting diodes (LEDs) illuminate the microrobots, one for powering each robot's on-board photovoltaics and one for communication. To transmit data, the communication LED sends an initial sequence of illumination flashes, called a passcode, that instructs the robot to receive subsequent flashes as bits and write them to its instruction memory bank (Fig. 1E). The passcode prevents random fluctuations from altering the robot's state and each robot recognizes two: a global passcode common to all robots and a type-specific passcode that enables us to send different instructions to specific subsets of robots.

The process for generating optical instructions is fully automated through a graphical user interface and a custom-built printed circuit board that drives the communication LED, allowing virtually anyone to program the robot. Yet once instructions are written to the memory, the robot's behavior is fully autonomous: the robot computes its actions based on the program in its on-board memory and its own sensor data (Fig. 1B) without further user input.

***Propulsion:*** Though there are many mechanisms for motion at the microscale(*22*, *23*), compatibility with the on-board electronics requires low-current (<100nA) and low voltage (~0.1-1V) operation. To our knowledge, two actuators fit these constraints: surface electrochemical actuators (*24*, *25*) and electrokinetic propulsion (*26*). The latter is the subject of this paper while the former is a topic of ongoing work.

The governing mechanism for electrokinetic propulsion is discussed in detail in our prior work (*26*); we review it briefly here. As shown schematically in Extended Data fig. S1, the robot, which must be immersed in a fluid, oppositely biases metal electrodes, passing a current between them. Mobile ions surrounding the robot and nearby surfaces move in response to this field, dragging the fluid with them. This establishes a flow which in turn causes the robot to move at a speed proportional to the applied electric field (*26*). By manipulating the electric field through spatial patterns of active electrodes, the robot can travel in different directions and/or turn with a power expenditure of approximately 60 nW at this size scale.

Compared to alternatives, electrokinetic propulsion offers several advantages. First, it is straightforward to implement: the electrodes themselves are simple layers of platinum, requiring no moving parts or complex fabrication steps. Consequently, actuators are robust, lasting upwards of months, and fabrication can be done massively in parallel using fully lithographic patterning. Second, the electrical signal is a DC voltage, requiring no up-conversion or complex temporal sequencing (as would be needed for legs). This reduces the circuit overhead needed to control the robot's behavior to a bare minimum, leaving more space for sensing, memory, and computation.

To integrate these actuators and release the robot from the underlying silicon wafer, we carry out a series of lithographic, low-temperature back-end fabrication steps, as shown in Fig. 2. This protocol dramatically improves on prior work for microrobot release based on specialized silicon on insulator wafers(*14*, *27*) by generalizing to wafers built at arbitrary semiconductor foundries in arbitrary process nodes. Briefly, we add an oxide border around each microrobot body on the layout sent to the foundry to avoid the presence of metal structures typically placed for planarity concerns in commercial semiconductor processes. Following full assembly of the circuits, we thin the backside of each chip to 50 micrometers using a combination of mechanical and plasma etching. Then, we deposit and pattern chrome layers on the top and bottom of the chip to act as masking layers and etch barriers, respectively. From the top, we etch through the oxide border to the underlying silicon with an inductively coupled plasma followed by etching through the remaining silicon wafer with deep reactive ion etching until hitting the bottom layer of chrome. This stopping layer both arrests the etch and supports the released robots. Finally, we wet etch the residual chrome, releasing the robots into solution. Typical yields exceed 50%.

Released chips with embedded actuators form microscopic robots, each with tightly integrated systems to sense, compute, communicate, and move. As a first test, we characterize robot locomotion with simple programs that directly set actuator polarity, and thus propulsion direction. Under digital control, a four-electrode robot has fourteen unique configurations that generate motion (2 of the 16 polarity states are degenerate, corresponding to all electrodes at the same voltage). As shown in Fig. 3, we find that the fourteen states can be grouped into four behaviors: translations along the major or minor axis, rotations, and arcs. Each of these groups is determined by the number and arrangement of positive polarity electrodes on the robot, with robots differing only by rotational symmetry within the same group. Statistics collected over 56 experimental trials show typical speeds range from 3-5um/sec for translations and 0.1-0.3°/sec for turns, with a statistical variability that depends on imposed behavior (Fig 3B). Further discussion of the effects of solution conductivity, polarity, current and expected improvements in speed via advanced circuitry are discussed further in the Supplementary Text.

Arcs, turns, and translations offer full control over the robot's three degrees of freedom, and thus can be strung together to trace out arbitrary, user-specified paths in a plane. Movies S1-3 show a base-station transmitting sequences of actuator commands to a robot and the robot updating its motion as it receives each message. Specific paths can also be given to specific robots by prefacing different passcodes to the instructions: Fig. S2 and Movie S4 shows two robots executing different sequences of motions, each responding only to its own, type-specific set of instructions.

***Sensing, Feedback, and Control:*** We can go directly from motion primitives to behaviors by implementing on-robot programs that transition between locomotion modes in response to sensory data. Here we explore two different closed loop behaviors with a robot adapting to its environment by measuring temperature. In one case, the robot responds to a changing temporal pattern and in the other it climbs spatial gradients.

For the temporal task, we expose the microrobots to a heating environment and program them to report back the local temperature by modulating their motion. Specifically, each microrobot measures temperature with its on-board sensors, then digitizes the value and transmits it back to a base station by switching the polarity of the front-right actuators with a Manchester encoded sequence (Fig. 4A). The resulting body movement is then decoded to read out the robot's temperature measurement in bits (Fig. 4B). Thanks to our custom-ISA, most of this program can be implemented with just two commands: 'ts' handles the sensing and 'wav' encodes and transmits the data. We validate the program by placing robots in a bath of cooled solution left to passively warm to room temperature. When we compare the robot's reported temperature measurements to those made simultaneously with a thermocouple, we find the agreement as shown in Fig. 4C.

Compared to other small digital sensing schemes, microrobots reporting temperature through motion are highly accurate (*28–42*). Figure 4D plots sensor accuracy against volume for a variety of state-of-the-art digital temperature measurement systems. Our robot pareto dominates the field, offering a 0.2°C resolution in a sub cubic mm volume. The handful of sensors that achieve comparable or better accuracy occupy at least an order of magnitude more space.

To show adaptation in response to spatially varying environments, we program microrobots to climb thermal gradients (i.e., thermotaxis). Using thermoelectric heat pumps, we introduce a temperature gradient in the solution bath. The robot is programmed to execute an arcing motion and find warmer regions if the current temperature reading is lower than the previous one. Alternatively, if the temperature value is warmer, the robot switches to a turning state, pivoting without translation. We choose these two states as they present a clear signal to validate the behavior and enable the robot to broadly explore space. Figure 4D and movie S5 show the experimental results. Initially, there is no imposed gradient, and the robot correctly turns in place. When the heat pump is turned on to cool the local area, the robot switches to an arcing state and explores the workspace until finding a warmer region, where it switches back to rotating. To further validate the program, we switch the direction of the temperature gradient. In response, the robot transitions back to arcing, now traveling in the opposite direction.

The combined capacity to explore space and report values at high precision makes these robots uniquely suited for precision temperature measurements in life sciences (*43, 44*). To monitor cellular processes, temperature reporting requires a small sensor volume and sub-degree temperature accuracy. In addition, any biological system is subject to change as cells grow, die, and restructure, necessitating a sensing platform that can suitably reposition itself over time. Our robots are the first technology to satisfy these demands. They meet or exceed other digital measurement schemes in terms of precision and size and can autonomously adapt their own placement, finding interesting or extremal temperatures, before reporting measurements. Interfacing robots with biology for temperature reporting would also be straightforward. Since heat flows through materials, the robots in this work could be used as is by placing their aqueous operating environment above the targeted substrate and monitoring robot actions after the robot and biological media have reached thermal equilibrium. Keeping the environments separate but thermally coupled would enable accurate measurements while bypassing the need for biocompatibility.

**Discussion:** Many other microrobots have achieved taxis-like behaviors (*45, 46*), often using vastly simpler analog effects. Such devices represent an elegant counterpoint to our approach. Rather than fabricating sophisticated circuits, one can embed responsivity in a material to achieve similar goals (*12*) with significantly lower manufacturing and design complexity.

However full, programmable autonomy with local sensor feedback is a fundamentally new capability for microscopic robots that brings broad, distinctive benefits. First, many current microrobot platforms require large pieces of specialized laboratory equipment like magnetic coils (*47*), ultrasound transducer arrays (*48*), or cell culture environments (*49*). This overhead restricts use to experts with bespoke, expensive equipment, even if the cost per robot is low. By contrast, we require only a controllable light source for power and programming because control has been offloaded to the robot. The result is high-level functionality using only commonplace parts. Second, current microrobots that can sense or control their behavior execute a fixed, design-specific set of tasks (*12*, *13*). In contrast, our robots integrate a complete, fully functional computer in a few hundred microns of space. As shown, digital programming and on-board computing allow a single, general-purpose microrobot to carry out a range of tasks which can be reconfigured on demand, post-fabrication.

This level of flexibility invites new applications where the state of the environment, task, and/or the robot itself are unknown or subject to change. For instance, the capacity to sense and respond directly to local sensor data (e.g., biochemical markers or hyper/hypo-thermic conditions) could enable microrobots to increase their accuracy in targeted drug delivery applications, releasing drugs based on local cues instead of global guidance. Similarly, robots used in telemetry applications (e.g., monitoring body temperature) could greatly improve their accuracy by using on-board computation to digitally encode data before transmission (e.g., Manchester encoding), thereby rendering the communication channel robust to noise. Finally, microrobots in nanomanufacturing could construct more complex materials or structures by leveraging programmability and communication: using passcode-based communication, robots could be given independent instructions, monitored, and updated as they complete objectives.

Progress towards these applications, while long-term goals, could largely be achieved by further integration of semiconductor technology. For example, we estimate that the ~kbits of memory on each robot here could be increased nearly 100-fold by changing to a more advanced process node, provided the threshold voltage for circuit operation remains high enough to mitigate leakage. Such an expansion would allow for significantly more complex programs to run (i.e., upwards of a few thousand lines), and by extension more sophisticated autonomous behaviors. Following the work here, programs can be made unique to specific robots via the already integrated passcodes. In the absence of direct inter-robot communication, this allows a central controller to send instructions and/or emulate communication and interactions between agents. Furthermore, by better tailoring the actuator subcircuits to specific propulsion technologies, we estimate the speed of the robot could be enhanced by as much as 10x (see Methods), enabling robots that move at speeds of a few body lengths per second under on-board control. Future circuits could also add new functionalities like localization of robots in specific structures, communication through the surrounding media, or coordination across long length scales.

Remarkably, these benefits could be achieved without incurring lab overhead or dramatically increasing the cost per machine, which currently stands at roughly a penny per robot (see methods for details of calculation). The combination of versatility, ease of use, and low-cost offered by on-board information processing could impact a range of communities, enabling new approaches for studying the physics of living systems(*50*, *51*)**,** nanomanufacturing(*52*), microsurgery(*23*, *53*), or drug delivery (*54*). Indeed, every living cell speaks to the broad promise of intelligent robotic systems too small to see by the naked eye.

## Materials and Methods

Fabrication and Release:

We start with our custom processor fabricated in commercially produced 55nm CMOS chips (Fujitsu). Before fabrication, the chip is thinned to 50 microns using mechanical and plasma etching. The chip is uniformly passivated with 300nm of silicon oxide via plasma-enhanced chemical vapor deposition using silane. After passivation, we etch holes through the passivating oxide using reactive ion etching with fluorine-based chemistry to expose the actuator control pads. Platinum electrodes are then sputtered onto these vias and patterned with liftoff in a heated sonication bath. Uniform chromium is deposited onto both sides of the chip and the top surface is patterned with wet chromium etchant 1020AC to expose the dielectric border surrounding each robot body. The dielectric border is etched in a CHF3 inductively coupled plasma etch. After the dielectric border is etched, the bulk silicon surrounding each robot is removed using the Bosch process, stopping on the backside uniform chromium. Finally, the protective chromium is removed in wet chromium etchant 1020AC and the robots freely release from the array.

Using Robots after Release:

In our prior work, we found electrokinetic propulsion could be achieved in a variety of different solution environments, ranging from DI water to weak acid/base solutions, and on a range of substrates, including glass, metal, and polymers. The principal role of the solution environment is to set the effective mobility of the robot (i.e., the proportionality between applied field and robot speed). While this effect is minor, the highest mobility identified thus far differs only from the lowest by a factor of roughly 4, we find experimentally that working in 5mM hydrogen peroxide provides one of the largest mobilities with a wide experimental working window. To prepare the bath, 25μL of 30% H2O2 is added to 50mL of DI, achieving a 5mM concentration, followed by the robots for testing. Experiments were carried out in polystyrene petri dishes. Unless purposefully reprogrammed in groups (such as in movie S4), robots are tested individually. When not in use, robots are kept in DI baths and stored in a 4°C refrigerator away from light. Robots reliably reprogram with proper illumination conditions and storage for several months.

Illumination Conditions for Operating Robots:

The optical receiver onboard, based on silicon photovoltaics, is sensitive to a range of different wavelengths. Thus, we use two separate wavelengths for power and communication, filtering the modulating communication wavelength out when imaging. Here, the power LED has a center wavelength of 473nm and the communication LED, 565nm. We find the base incident illumination intensity must exceed 200 $W/m^2$ to support sufficient power for onboard processing and locomotion. The maximum power tolerated by the optical receiver sets the intensity ceiling for the robots, measured to be 4000 $W/m^2$. These values were measured with a Thorlabs Optical Power Meter (PM100D). We note a standard operating illumination of 600 W/m² power with the communication channel contributing 1000 W/m² to reach a peak incident power of 1600 W/m². Various combinations of base power and communication channel intensities were tested, shown as a plot in figure S5. We note that this operation is near the typical solar power of 1300 W/m² on a sunny day *(55)*.

Characterization of Photovoltaics and Detailed Power Budget:
We measured the solar cell I-V characteristic by reprogramming to the DC forward drive state; the actuator subcircuits are routed to the positive and negative terminals of a PV (Fig. S3). We find incident optical power goes through a 10x reduction, typical for the photovoltaic conversion in silicon, and measure a responsivity of 0.34 A/W.

To maximize available surface area for solar cells, the underlying circuitry for computation is optimized for area, such as selecting a subthreshold temperature sensor(*56*), adopting a leakage-based optical receiver(*57*). To address power, it is necessary to use older technologies that reduce leakage however this comes with the consequence of larger logic cells, creating a tradeoff between leakage and intelligence. To minimize the silicon area of the processor, we use a custom designed, compact 11-bit instruction-set including arithmetic calculations, reading, and writing memory. This instruction-set is general enough to support arbitrary programs with complex control flow while it uses only 11 bits per instruction minimizing code size and, therefore, memory footprint. In addition, the instruction set supports several instructions targeted specifically at the operation of the robot, such as fetching temperature sensor data, and operating the actuation. This allows the program to operate the temperature sensor and initiate a particular motion such as turn left or right with a single instruction. This greatly reduces the required memory size while increasing combinational logic in the processor only marginally. Table 1 details the instruction-set codes and functionality in full.

Detailed Description of Communication Protocol:
Users send instructions with a python graphical user interface (GUI) that has physical descriptions of robot behaviors. The python GUI compiles these robot behaviors into the assembly language of the onboard computer and, finally, into a binary bit stream. This bit stream pulses a communication LED (comms LED) through a Raspberry Pi Pico mediated TTL connection at a user-set frequency. The optical receiver onboard the robot locks into the comms LED to rewrite the memory stack and change behaviors while deployed in solution. The optical receiver is always on and monitors the incident light power to detect whether a passcode has been received. Each reprogramming sequence has instructions that are interpreted by the onboard computer and new behavior is executed by the robot. After reprogramming terminates, the robot may remain tied to the oscillating comms LED channel as the clock, useful for user defined paths and studying locomotion. Otherwise, the robot can be programmed to return to the onboard clock and completes tasks autonomously – without further user input. Whether the robot switching between these states is included as part of the transmitted programs.

Upon startup, only the power LED is illuminated, and the robot uses this baseline power to wake up a power-on detection circuit and enter the default state – a sequence of 32 oscillations between front and rear electrodes followed by a ~10 second pause period before repeating this behavior. This default behavior uses the onboard clock and is temperature sensitive, with heat reducing the total loop time from 60 to 20 seconds for a 10°C to 40°C. At room temperature the default cycle takes approximately 30 seconds. We use this state to validate the robot is viable before carrying out experiments.

Reprogramming has an upper constraint on the light level to avoid saturation of the optical receiver. Thus, we bound the power LED intensity to ensure the comms LED pulses remain below the optical receiver maximum. The optical intensity range for reprogramming is found by experimentally measuring the delivered optical power (Thorlabs PM100D) then confirming successful reprogramming both on-chip via high impedance probe measurements, and by successfully sending new instructions to deployed swimming robots that can be monitored with object detection software. Acceptable intensities that result in successful reprogramming have a range of 20-2000 mW/mm$^2$, marking a 100x range and highlighting the ease of the reprogramming setup.

Detailed Description of Locomotion Data Collection:
For all of the propulsion modes in figure 3A, robots were monitored over roughly half hour intervals following reprogramming into each DC state. Center of mass and orientation data were extracted using image processing software (ImageJ) and linear regression was used to determine the rate of rotation and linear translation speeds. Experiments were repeated to produce >10 trials for each of the robot's behaviors (arcs, turns, translations) and the summary statistics in figure 3B. We note that, accounting for the different units in angular and linear velocity, the variance across data sets is consistent with a typical uncertainty in velocity on the order of 1 micron per minute.

Improving Locomotion via Circuits for Electrokinetic Propulsion:
The order of magnitude for electrokinetic propulsion speeds is set by the applied electric field in the solution. In other words, the robot's speed scales in proportion to the applied current output by the robot's electrodes and inversely proportional to the solution conductivity, as shown in our prior work on robots of comparable size and in comparable solutions.

Thus, by simply increasing the robot's output current in the electrode terminals, we could dramatically increase propulsion speed. This could readily be done in subsequent designs by increasing the operating voltage applied across the electrodes from its current value (~1V) to a value closer to the voltage at which hydrolysis takes place (~2V electrode to electrode). At this value, the current scales exponentially with further increases in voltage, resulting in the downstream power electronics seeing an effectively 'short circuit' and a robot can directly control its speed by simply modulating the current limits while holding the voltage nearly constant.

In our prior work, we found that robots operating near the water-window voltage could maintain speeds of nearly 1mm/second by applying currents on the order of 10-100 nA. Such a range should be attainable with our current photovoltaic supply given we already expend approximately 60 nA for actuation. Indeed, the primary limitation of the current circuits is that because the operating voltage is too low, the circuits to see a high impedance across each electrode which limits the achievable propulsive field.

Finally, we note that robot locomotion in our current design does not respect parity inversion of the applied voltage. That is, robots display different behaviors depending on the number of positively biased electrodes. We attribute this fact to different Faradaic reaction rates on the platinum electrodes for forward and reverse bias configurations. Specifically, we directly measure an asymmetry between the forward and reverse solution impedance via a cyclic voltammogram (sweep rate of 10mV/s) in Fig. S4. Since it is the field within the solution, not across the electrode, that drives propulsion, this asymmetry can lead to different propulsion states with more current flowing (and thus faster fluid flows) in states that wire more electrodes in the positive direction. This effect could also be mitigated by increasing the operating voltage to enable direct control of the electrochemical current (instead of the current configuration, which essentially controls voltage, and thus is sensitive to impedance mismatch at the electrodes).

Cost of a robot:
Robots in this work were purchased from a multi-project wafer run at a price on the order of $10,000 USD per $mm^2$ of area from a semiconductor vendor. This purchase results in roughly 100 chips, each of which contains over 100 robots (see Muse Semiconductor for up-to-date pricing). Thus, a conservative estimate of price would be under $1 per robot. If scaled up to production, the cost would continue to decrease as subsequent runs see a discounted price thanks to previous investment in the photolithographic masks and the opportunity to purchase a full wafer of robots. Both effects substantively lower the cost per area, making a $0.01 price per robot a realistic estimate of the robot as a commercial product.

**Acknowledgments:** The authors would like to thank Michael Reynolds and Professor Daniel Koditschek for helpful discussions. The authors would like to thank Mohsen Azadi and the Singh Center for Nanotechnology staff for technical support during the fabrication process.

**Funding:**

National Science Foundation grant NSF 2221576

the University of Pennsylvania Office of the President

Air Force Office of Scientific Research grant AFOSR FA9550-21-1-0313

Army Research Office grant ARO YIP W911NF-17-S-0002

the Packard Foundation

the Sloan Foundation

NSF National Nanotechnology Coordinated Infrastructure Program grant NNCI-2025608 which supports the Singh Center for Nanotechnology

**Author contributions:**
- Conceptualization: DB, MZM
- Methodology: MML, JL, LH, WHR, LX, MZM, DB, DS
- Investigation: MML, JL, KS
- Visualization: MML, JL
- Funding acquisition: DB, MZM, MY
- Project administration: MZM, DB
- Supervision: MZM, DB, MY
- Writing – original draft: MML, MZM
- Writing – review & editing: MML, MZM, JL, DB, KS, MY

**Competing interests:** The authors declare no competing interests.

**Data and materials availability:** All data are available in the main text or the supplementary materials. Supplementary information is available for this paper. Correspondence and requests for materials should be addressed to M.Z.M.

**Supplementary Materials**

Figs. S1 to S4

Table S1

Movies S1 to S4

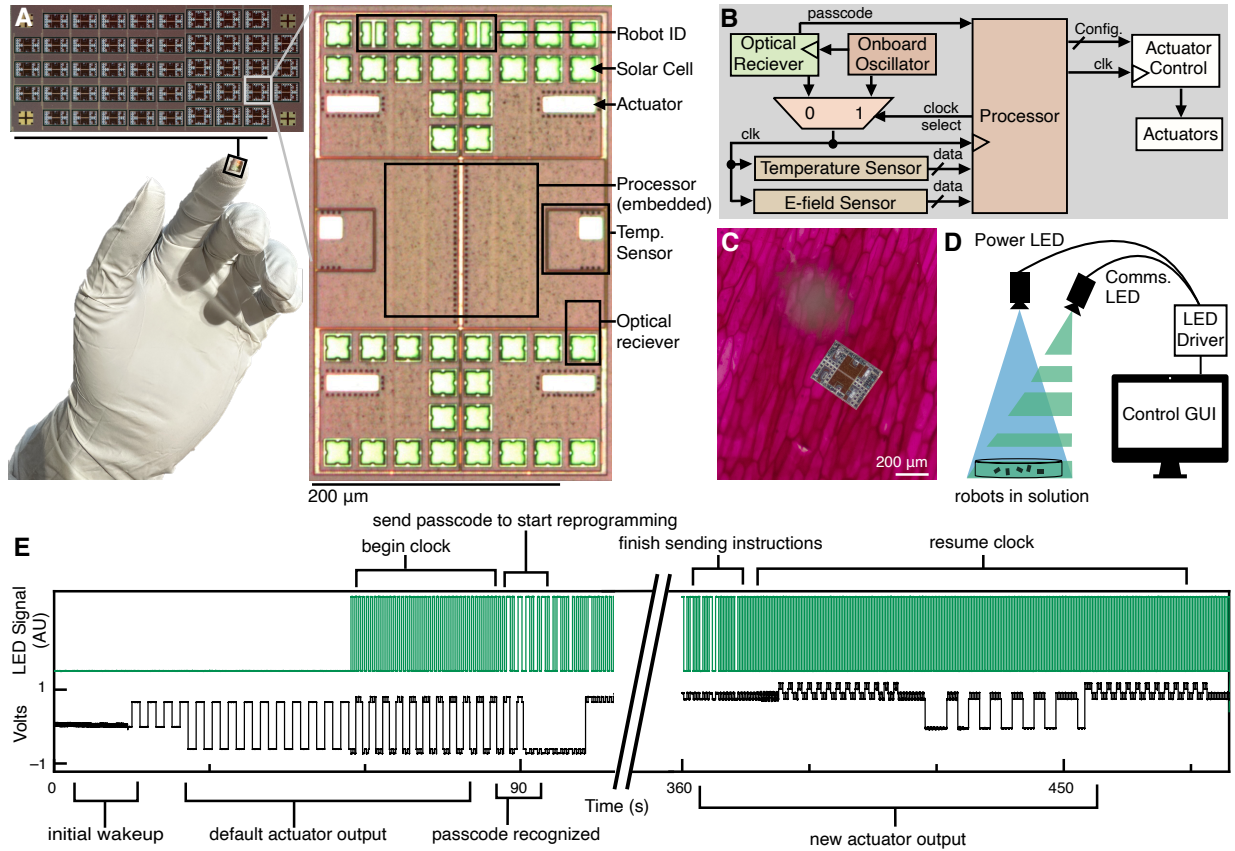

**Figure 1: Overview of the microrobot circuits.** A: Each robot contains several integrated pieces of microelectronics, spanning sensing, memory, processing, communication, and power. These devices are fabricated together in a 55 nm CMOS process at a commercial foundry and have been optimized for size and power. Data flows between the subsystems as depicted in B: Incoming information from the sensors and communication system are used by the processor to update actuator outputs. C: Composite photograph of a robot on top of plant cells. D: The robot is powered and programmed using light. To send instructions, an automated compiler modulates the optical power of a communication LED, with each flash encoding a bit in the program. E: To prevent random light fluctuations from interfering with the robot's operation, the communication protocol features a passcode sequence which robot must see to accept an incoming program, store the data to its memory, and execute.

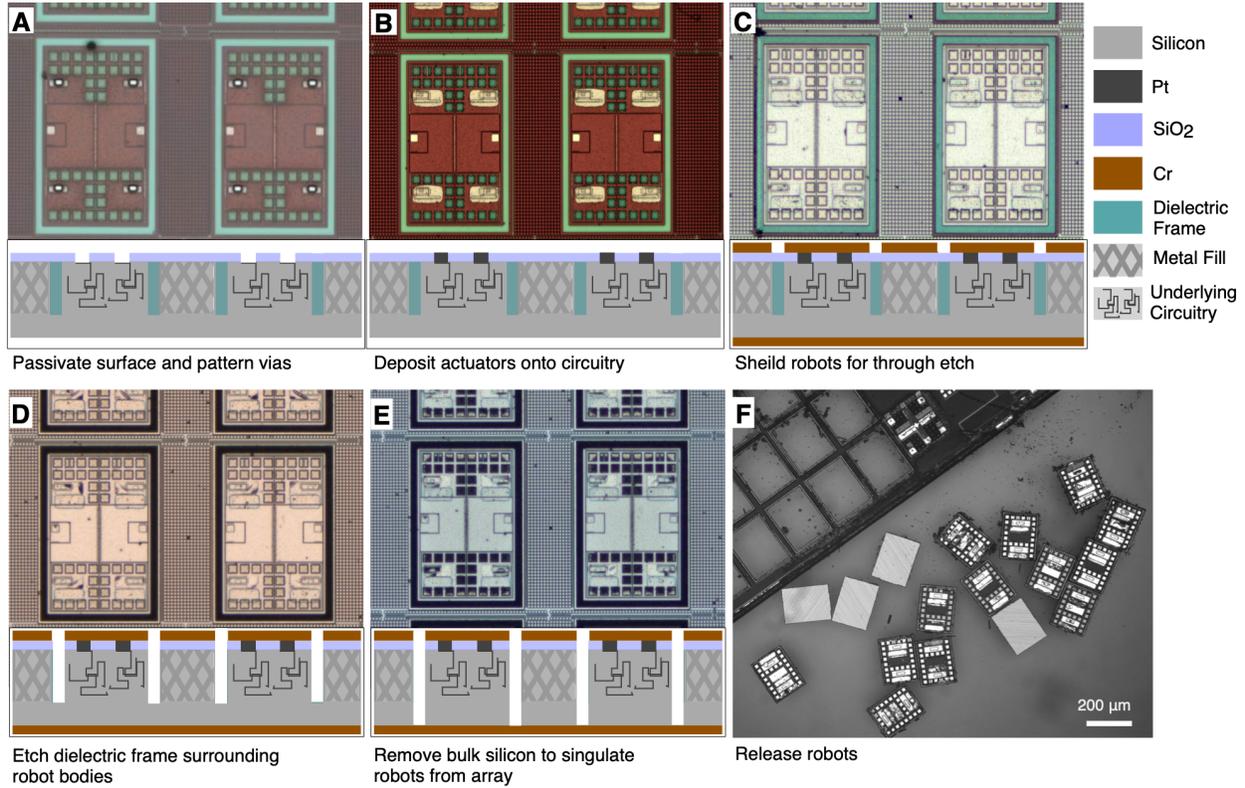

**Figure 2: Integrated fabrication and Release.** We transform foundry fabricated wafers into robots using six key steps. A: we passivate the surface using oxide layers and pattern metal interconnects to the underlying electronics. B: The actuators are deposited and then wired to the electronics through the interconnects. C: The remaining steps release the robots, shielding them with a metal hard mask, D: etching through the oxide layers of the wafer, and E: deep etching the underlying silicon wafer to free devices. F: When finished, the hard mask layer is removed, releasing robots into solution *en masse*. Typical yields are roughly 50%.

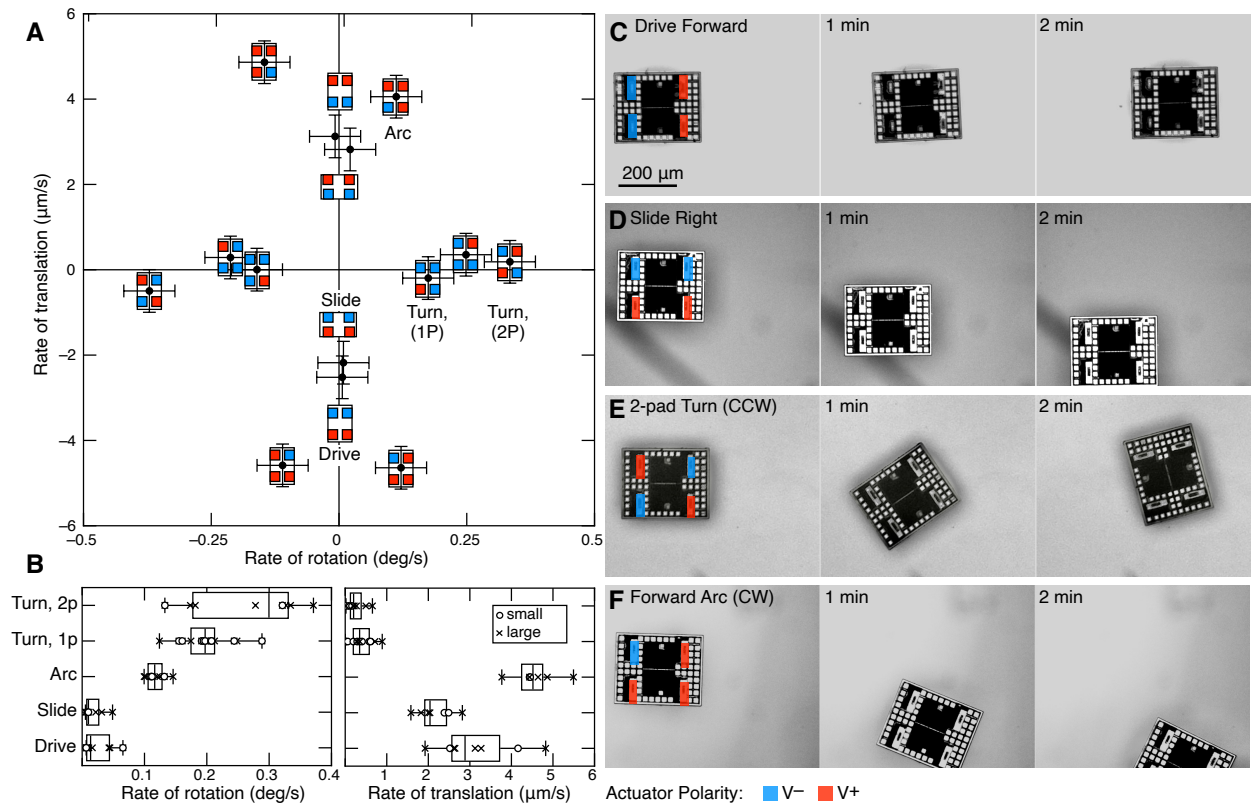

Figure 3: **Reprogrammable locomotion.** A: the robot's degrees of kinematic freedom, namely its signed rate of rotation and forward velocity, depend on the electric field produced in solution by passing current through its electrodes. By setting different electrode polarities, we can generate four classes of robot behaviors: translations along the major and minor axis, turns, and a superposition of translation with rotation, here called arcs. The robot's motions generally respect symmetries of the robot body, with states differing only by rotations or reflections matching in terms of velocity and rotation magnitude. A: Data points correspond to individual experiments with error bars set by estimates of the microscopy drift and ambient fluid flows. B: Detailed statistics of robot motion show the device-to-device variation over an ensemble of 56 experiments. Data collected for the two robot sizes shows no significant difference between the large and small robot designs. C-F: A time lapses the four representative body motions over 2 minutes.

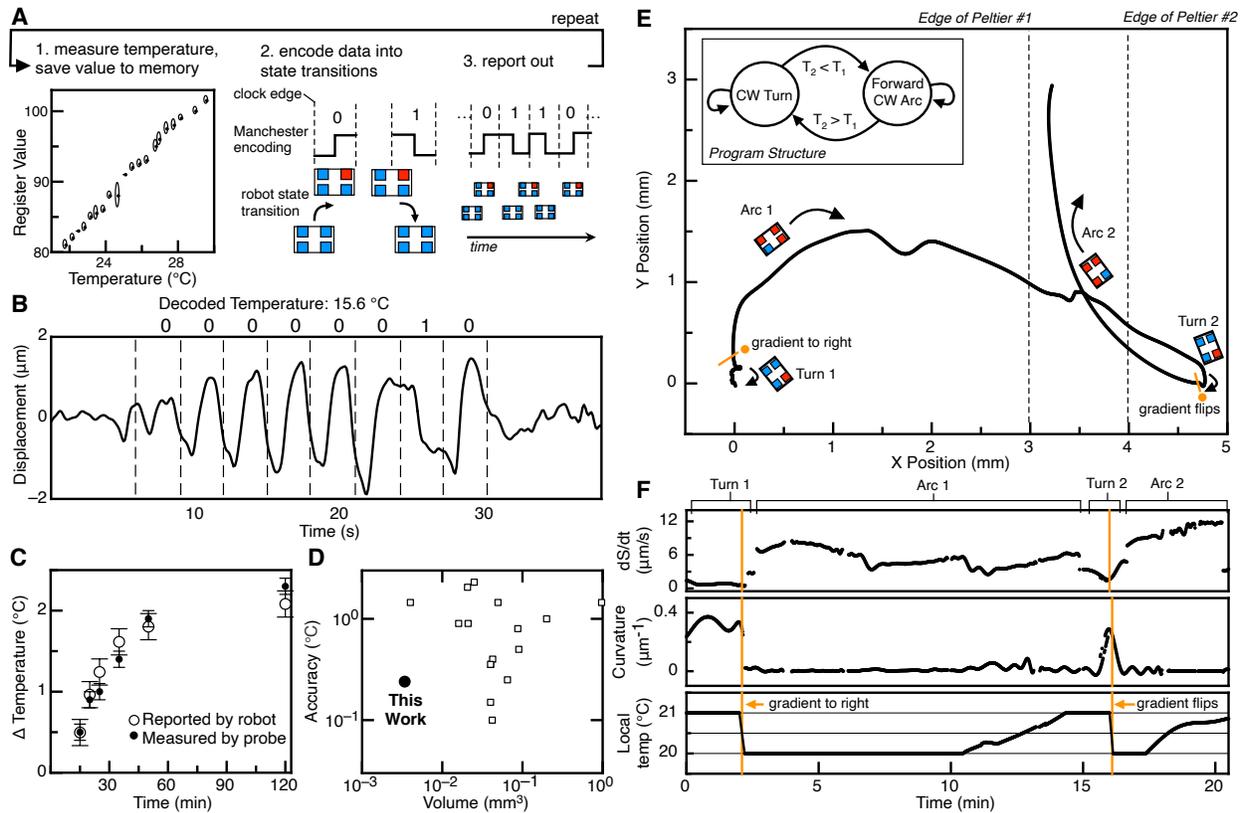

**Figure 4: Motion controlled by sensor feedback.** A: A program where the microrobot responds to temporal changes in temperature, periodically measuring the environmental temperature, storing the measurement to memory, and modulating its actuators to report data back the data using Manchester encoding. B: The resulting displacements can be decoded to give a measurement of the local temperature. C: By placing the robot in a warming bath, we can compare its measurements to those made with a probe, finding agreement. D: We plot temperature accuracy and total volume for our results and existing works, references 28-41, individual system details available in table S1. E: A program where robots respond to spatial variations in temperature. Here a robot is programmed to executing an arcing motion whenever the temperature drops below a previously measured value (inset), otherwise it turns in place. In this way, robots can either explore the environment or hold a position, depending on their local sensor readings. When we turn on a thermal gradient, we find robots execute the arc until they reach a cooler zone, at which point they begin turning. Reversing the gradient direction triggers the robot back into the arcing state. F: Curvature, speed, and estimated local robot temperature from thermal imaging confirm the transitions take place in response to new, local temperature measurements.

| Category | Instruction | Operation | Meaning |
|---|---|---|---|
| Actuator | mot | Actuator motion (N) | Enable select actuators, then repeat programmed motion N times |
| | wav | Mx → Actuator motion | Modulate select actuator motion by Manchester-encoded pattern of data in Mx |
| Sensing | ts | Temperature → Mx | Sense temperature, then store data in Mx |
| Data transfer | mov | Rx → Ry | Move data from Rx to Ry |
| | sb | Rx → Mx | Store byte in Rx to Mx |
| | lb | Mx → Rx | Load byte in Mx to Rx |
| | li | Imm → Rx | Load Imm to Rx |
| Arithmetic | add | Rx + Ry → Rz | Add Rx and Ry data, then store the result in Rz |
| | addi | Rx + Imm → Rz | Add Rx data and Imm, then store the result in Rz |
| | sub | Rx - Ry → Rz | Subtract Ry data from Rx data, then store the result in Rz |
| | subi | Rx - Imm → Rz | Subtract Imm from Rx data, then store the result in Rz |
| | and | Rx & Ry → Rz | Bitwise AND for Rx and Ry data, then store the result in Rz |
| | or | Rx \| Ry → Rz | Bitwise OR for Rx and Ry data, then store the result in Rz |
| | sll | Rx << Imm → Rx | Shift left logically for Rx data of the amount of Imm, then store the result in Rx |
| | srl | Rx >> Imm → Rx | Shift right logically for Rx data of the amount of Imm, then store the result in Rx |
| Unconditional jump | j | go to A | Jump to address A |
| | bcnt | go to -O (N) | Jump to address with offset -O, then repeat the in-between instructions N times |
| Conditional branch | cmp | compare Rx, Ry | Compare Rx and Ry data, then set flag bits, eq/ne/gt/lt |
| | cmpi | compare Rx, Imm | Compare Rx data and Imm, then set flag bits, eq/ne/gt/lt |
| | beq | go to +O if equal | Conditional branch on eq to address with offset O |
| | bne | go to +O if not equal | Conditional branch on ne to address with offset O |
| | bgt | go to +O if greater than | Conditional branch on gt to address with offset O |
| | blt | go to +O if less than | Conditional branch on lt to address with offset O |

\* Rx, Ry, Rz: arbitrary register file addresses
\* Imm: immediate value
\* Mx: arbitrary memory address
\* A: arbitrary instruction address
\* O: arbitrary address offset relative to the current program counter

**Table 1: Custom-designed 11-bit instruction set for onboard processor.**

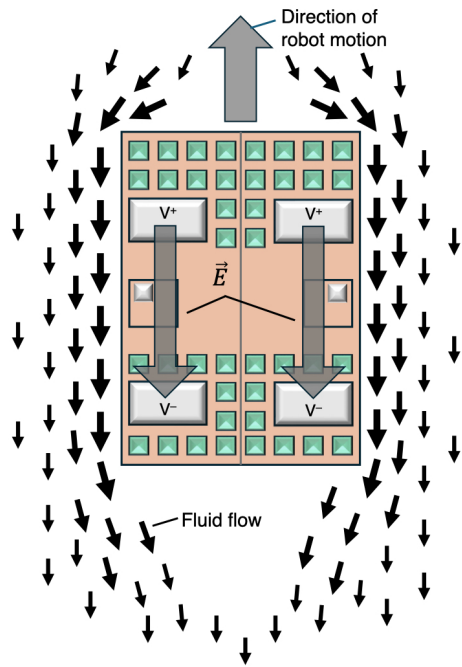

**Fig. S1.**

Schematic depicting the fluid flow surrounding the robot when in the forward drive state. A static electric field is established between front and rear actuators with resulting fluid flow and consequent forward translation.

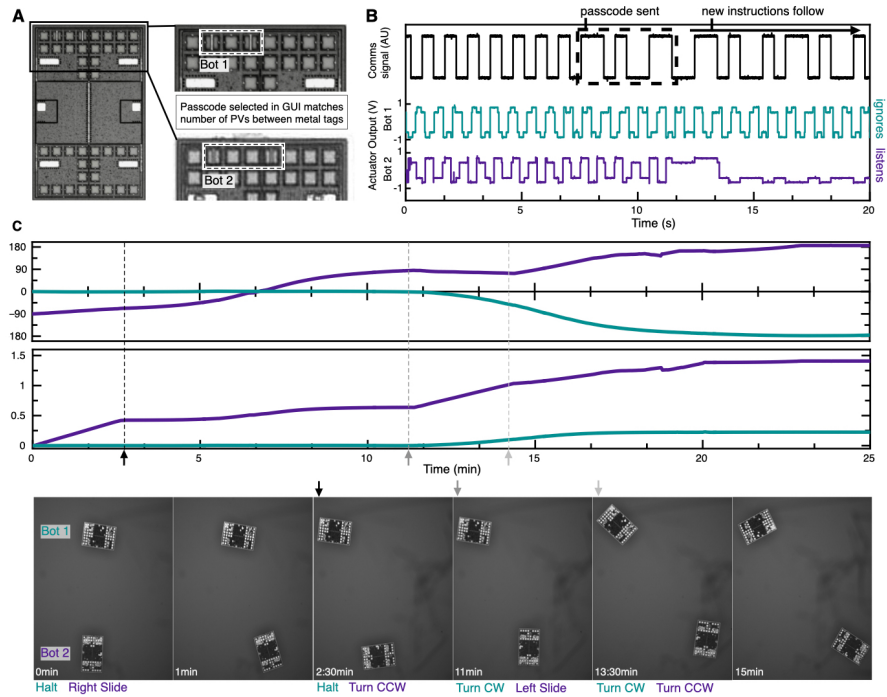

**Fig. S2.**

A, Robot ID selected during reprogramming matches the number of solar cells between vertical metal tags and values stored in memory. B, A plot of the Comms LED sending a Robot 2 specific reprogramming sequence. Robot 1 is unaffected and ignores the incoming instructions. Robot 2 stops previous task and listens for new instructions to follow.

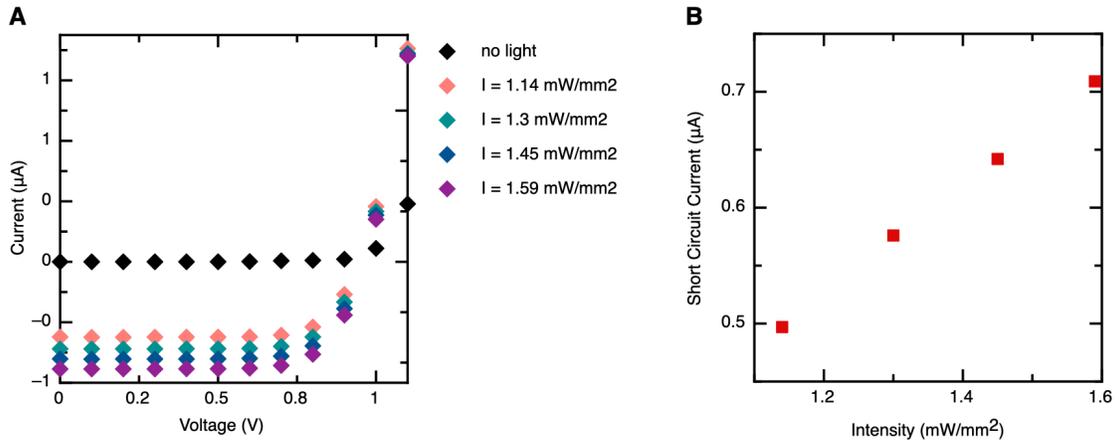

**Fig. S3.**

A: experimental data measuring the current-voltage characteristic between the front and rear actuators when the robot is programmed into the forward drive state across varying intensities. B: resulting photovoltaic responsivity.

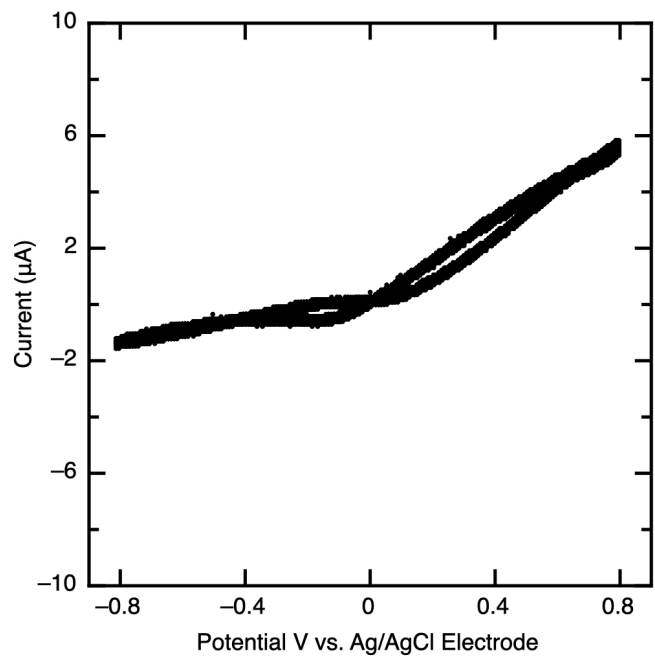

**Fig. S4.**

Cyclic voltammogram performed between a platinum electrode and an Ag/AgCl counter electrode in 5mM H2O2 solution. Scan rate of 10mV/s.

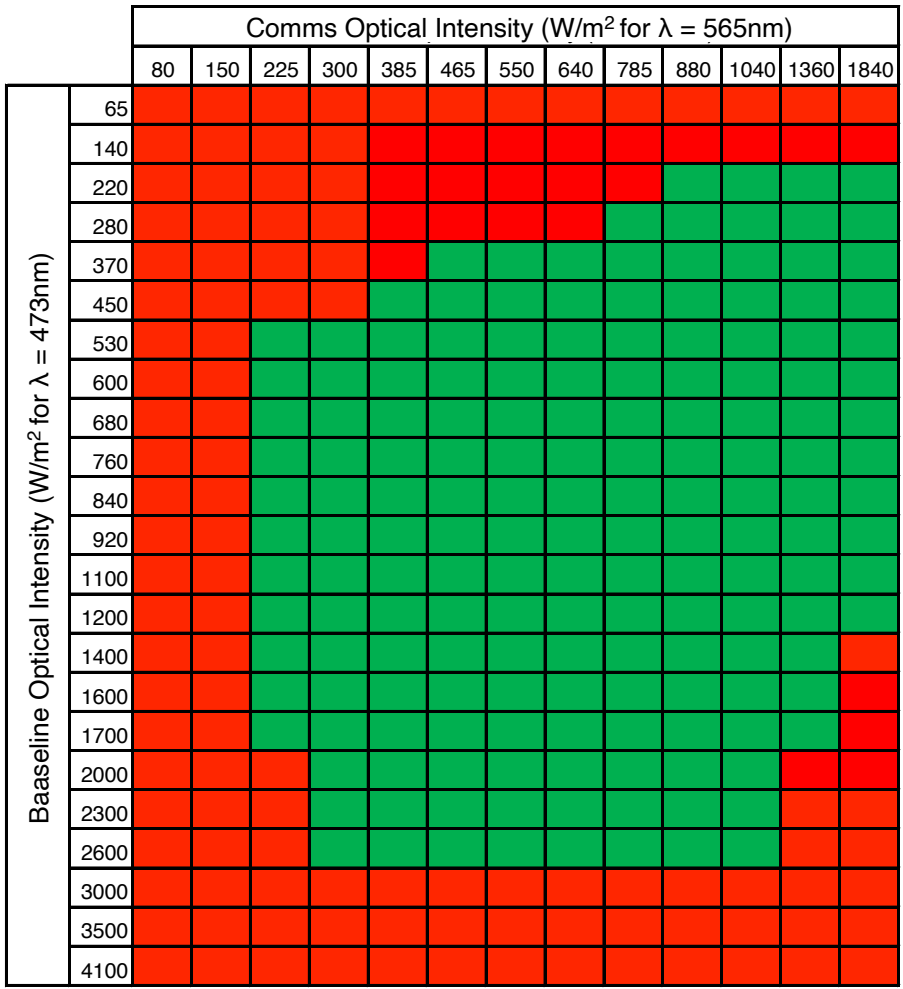

**Fig. S5.**
Plot characterizing robot illumination and reprogramming. Green denotes successful reprogramming sequences; red denotes failed reprogramming attempts.

| ref | vol (mm³) | precision + °C | precision - °C | power (nW) | average accuracy |
|---|---|---|---|---|---|
| 28 | 0.065 | 0.22 | 0.28 | 0.8 | 0.25 |
| 29 | 0.04 | 0.38 | 0.33 | 16 | 0.355 |
| 30 | 0.09 | 1.5 | 1.4 | 71 | 1.45 |
| 31 | 0.025 | 3 | 1.6 | 220 | 2.3 |
| 32 | 0.042 | 0.1 | 0.1 | 110 | 0.1 |
| 33 | 0.021 | 1 | 0.8 | 120 | 0.9 |
| 34 | 0.016 | 1 | 0.8 | 405 | 0.9 |
| 35 | 0.0425 | 0.4 | 0.4 | 600 | 0.4 |
| 36 | 0.0875 | 0.9 | 0.7 | 10000 | 0.8 |
| 37 | 0.2 | 1 | 1 | 9000 | 1 |
| 38 | 0.02065 | 2.7 | 1.4 | 23100 | 2.05 |
| 39 | 0.09 | 0.5 | 0.5 | 24000 | 0.5 |
| 40 | $4 \times 10^{-3}$ | 1.5 | 1.4 | 500000 | 1.45 |
| 41 | 0.04 | 0.15 | 0.15 | 5100 | 0.15 |
| This Work | $3.4 \times 10^{-3}$ | 0.24 | 0.24 | 100 | 0.24 |

**Table S1.**
System performance comparison for peer works plotted in Figure 4D.

**Movie S1.**
Large robot performing a sequence to run and then pivot clockwise. (The video has been sped up by 10x).

**Movie S2.**
Large robot performing a sequence of slides and turns, camera moves to follow robot progress. (The video has been sped up by 20x).

**Movie S3.**
Large robot performing a sequence of turns, reprogramming signal is filtered out. (The video has been sped up by 10x).

**Movie S4.**
Two medium robots following a series of ID specific instructions, resulting in two different paths. (The video has been sped up by 10x).

**Movie S5.**
Large robot changing locomotion modes in response to temperature sensor feedback. The gradient is imposed by a Peltier pump held below the petri dish. (The video has been sped up by 15x).